\documentclass{article}

\pdfoutput=1 

\usepackage{arxiv}

\usepackage[utf8]{inputenc} 
\usepackage[T1]{fontenc}    
\usepackage{hyperref}       
\usepackage{url}            
\usepackage{booktabs}       
\usepackage{amsfonts}       
\usepackage{nicefrac}       
\usepackage{microtype}      
\usepackage{lipsum}		
\usepackage{graphicx}
\usepackage{natbib}
\usepackage{doi}

\usepackage{times}
\usepackage{latexsym}
\usepackage{booktabs}
\usepackage{amsmath}
\usepackage{graphicx,wrapfig}
\usepackage{float}
\usepackage[justification=centering]{caption}
\usepackage[dvipsnames]{xcolor}

\usepackage{enumitem}
\setenumerate{noitemsep,topsep=0pt,parsep=0pt,partopsep=0pt}

\usepackage{hyperref}
\hypersetup{
    colorlinks=true,
    linkcolor=blue,
    filecolor=magenta,      
    urlcolor=blue,
    citecolor=blue,
}
\restylefloat{table}

\title{A Survey of Text Games for Reinforcement Learning informed by Natural Language}

\date{September 2021}

\author{Philip Osborne, Heido Nõmm \& Andre Freitas \\
  University of Manchester UK \\
  Kilburn Building University of Manchester, \\
  Oxford Rd, Manchester M13 9PL \\
  \{philip.osborne, heido.nomm, andre.freitas\}@manchester.ac.uk
 }

\begin{document}
\maketitle
\begin{abstract}

Reinforcement Learning has shown success in a number of complex virtual environments. However, many challenges still exist towards solving problems with natural language as a core component. Interactive Fiction Games (or Text Games) are one such problem type that offer a set of partially observable environments where natural language is required as part of the reinforcement learning solutions. 

Therefore, this survey's aim is to assist in the development of new Text Game problem settings and solutions for Reinforcement Learning informed by natural language. Specifically, this survey summarises: 1) the challenges introduced in Text Game Reinforcement Learning problems, 2) the generation tools for evaluating Text Games and the subsequent environments generated and, 3) the agent architectures currently applied are compared to provide a systematic review of benchmark methodologies and opportunities for future researchers.


\end{abstract}

\section{Introduction}

\begin{wrapfigure}{r}{0.25\textwidth} \label{textgame-example}
    \centering
    \includegraphics[width=0.25\textwidth]{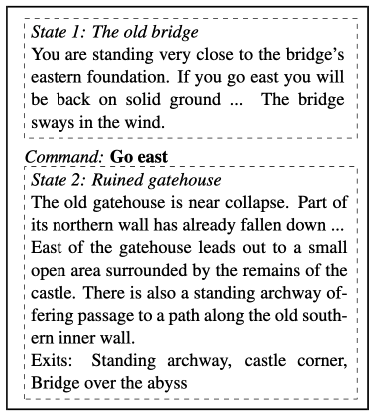}
    \caption{Sample gameplay from a fantasy Text Game as given by \cite{Narasimhan:2015} where the player takes the action `Go East' to cross the bridge.}
\end{wrapfigure}

Reinforcement Learning (RL) has shown human-level performance in solving complex, single setting virtual environments \cite{Mnih:2013} \& \cite{Silver:2016}. However, applications and theory in RL problems have been far less developed and it has been posed that this is due to a wide divide between the empirical methodology associated with virtual environments in RL research and the challenges associated with reality \cite{dulac:2019}. Simply put, Text Games provide a safe and data efficient way to learn from environments that mimic language found in real-world scenarios \cite{Shridhar:2020}.




Natural language (NL) has been introduced as a solution to many of the challenges in RL \cite{luketina:2019}, as NL can facilitate the transfer of abstract knowledge to downstream tasks. However, RL approaches on these language driven environments are still limited in their development and therefore a call has been made for an improvement on the evaluation settings where language is a first-class component.

Text Games gained wider acceptance as a testbed for NL research following work from \cite{Narasimhan:2015} who leveraged the Deep Q Network (DQN) framework for policy learning on a set of synthetic textual games. Text Games are both partially observable (as shown in Figure 1) and include outcomes that make reward signals simple to define, making them a suitable problem for Reinforcement Learning to solve. However, research so far has been performed independently, with many authors generating their own environments to evaluate their proposed architectures. This lack of uniformity in the environments makes comparisons between authors challenging and a need for structuring recent work is essential for systematic comparability.

Formally, this survey provides the first systematic review of the challenges posed by the generation tools designed for Text Game evaluation. Furthermore, we also provide details on the generated environments and resultant RL agent architectures used to solved them. This acts as a complement to prior studies in the challenges of real-world RL and RL informed by NL by \cite{dulac:2019} \& \cite{luketina:2019} respectively.

\section{Text Game Generation}

Thus far, there are two main generation tools for applying agents to interactive fiction games: TextWorld and Jericho.

\textbf{TextWorld} \cite{Cote:2018} is a logic engine to create new game worlds, populating them with objects and generating quests that define the goal states (and subsequent reward signals). It has since been used as the generation tool for Treasure Hunter \cite{Cote:2018}, Coin Collector \cite{Yuan:2018}, CookingWorld \cite{Trischler:2019} and the QAit Dataset \cite{Yuan:2019}.

\textbf{Jericho} \cite{Hausknecht:2019jericho} was more recently developed as a tool for supporting a set of human-made interactive fiction games that cover a range of genres. These include titles such as Zork and Hitchhiker's Guide to the Galaxy. Unsupported games can also be played through Jericho but will not have the point-based scoring system that defines the reward signals. Jericho has been as the generator tool for CALM \cite{Yao:2020} and Jericho QA \cite{Ammanabrolu:2020grue}. 


\subsection{Environment Model}

Reinforcement Learning is a framework that enables agents to reason about sequential decision making problems as an optimization process \cite{Sutton:1998}. Reinforcement Learning requires a problem to be formulated as a Markov Decision Process (MDP) defined by a tuple $< S, A, T, R, \gamma >$ where $S$ is the set of states, $A$ is the set of actions, $T$ is the transition probability function, $R$ the reward signal and $\gamma$ the discount factor. 

Given an environment defined by an MDP, the goal of an agent is to determine a policy $\pi(a|s)$ specifying the action to be taken in any state that maximizes the expected discounted cumulative return $\sum_{k=0}^{\infty}\gamma^k r_{k+1}$.

In text games however, the environment states are never directly observed but rather textual feedback is provided after entering a command. As specified by \cite{Cote:2018}, a text game is a discrete-time Partially Observed Markov Decision Process (POMDP) defined by $< S, A, T, \Omega, O, R, \gamma >$ where we now have the addition of the set of observations ($\Omega$) and a set of conditional observed probabilities $O$. Specifically, the function $O$ selects from the environment state what information to show to the agent given the command entered to produce each observation $o_t \in \Omega$.

\subsection{Handicaps} \label{handicaps}

Before introducing the challenges of the environments, it is important to understand the possible limitations that can be imposed by the generation tools to reduce the complexity of each problem. These handicaps are typically used to limit the scope of the challenges being faced at any one time for more rigorous comparisons and for simplicity. 

It has been noted that TextWorld's \cite{Cote:2018} generative functionalities explicit advantage in that it can be used to focus on desired subset of challenges. For example, the \textbf{size of the state space} can be controlled and how many commands are required in order to reach the goal. Evaluation of specific generalisability measures can also be improved by controlling the training vs testing variations. 

The \textbf{partial observability} of the state can also be controlled by augmenting the agent's observations. It is possible for the environment to provide all information about the current game state and therefore reduce the amount an agent must explore to determine the world, relationships and the objects contained within.  

Furthermore, the \textbf{complexity of the language} itself can be reduced by restricting the agent's vocabulary to in-game words only or the verbs to only those understood by the parser. The grammar can be further simplified by replacing object names with symbolic tokens. It is even possible for the language generation to be avoided completely by converting every generated game into a \textit{choice-based} game where actions at each timestep are defined by a list of pre-defined commands to choose from. 

\textbf{Rewards} can simplified with more immediate rewards during training based on the environment state transitions and the known ground truth winning policy rather than simply a sparse reward at the end of a quest as normally provided.

\textbf{Actions} are defined by text commands of at least one word. The interpreter can accept any sequence of characters but will only recognize a tiny subset and moreso only a fraction of these will change the state of the world. The action space is therefore enormous and so two simplifying assumptions are made:

	- \textit{Word-level Commands} are sequences of at most L words taken from a fixed vocabulary V.
	
	- \textit{Syntax Commands} have the following structure - verb[noun phrase [adverb phrase]] where [...] indicates that the sub-string is optional. 			
	

Jericho \cite{Hausknecht:2019jericho} similarly has a set of possible simplifying steps to reduce the environment's complexity. Most notably, each environment provides agents with the set of valid actions in each game's state. It achieves this by executing a candidate action and looking for the resulting changes to the world-object-tree. To further reduce the difficulty of the games, optional handicaps can be used:

\begin{itemize}
    \item Fixed random seed to enforce determinism
    \item Use of load, save functionality
    \item Use of game-specific templates and vocabulary
    \item Use of world object tree as an auxiliary state representation or method for detection player location and objects
    \item Use of world-change-detection to identify valid actions
\end{itemize}


Jericho also introduces a set of possible restrictions to the action-space.

\begin{itemize}
    \item \textit{Template-based} action spaces separate the problem of picking actions into two (i) picking a template (e.g. ``take \_\_\_ from \_\_\_"); (ii) filling the template with objects (e.g. ``apple", ``fridge"). Essentially this reduces the issue to verb selection and contextualised entity extraction.
    \item \textit{Parser-Based} action spaces require agent to generate a command word-per-word, sometimes following a pre-specified structures similar to ($\text{verb}, \text{object}\_1, \text{modifier}, \text{object}\_2$).
    \item \textit{Choice-based} requires agents to rank predefined set of actions without any option for ``creativity" from the model itself.

\end{itemize}

Lastly, the observation space may be enchanced with the outputs of bonus commands such as ``look" and ``inventory". These are commands agent can issue on its own but are not considered an actual step in the exploration process that could be costly and produce risks in the real-world to ask for. 


\subsection{Challenges and Posed Solutions}

Environments built from the TextWorld generative system bounds the complexity by the set number of objects available. For example, \cite{Cote:2018} introduce 10 objects including the logic rules for doors, containers and keys, where complexity can be increased by introducing more objects and rules into the generation process. The most challenging environments are defined in Jericho as these contain 57 real Interactive Fiction games which have been designed by humans, for humans. Specifically, these environments contain more complexity in forms of stochasticity, unnatural interactions, unknown objectives - difficulties originally created to trick and hamper players. 

The design and partially observed representation of text games creates a set of natural challenges related to reinforcement learning. Furthermore, a set of challenges specific to language understanding and noted by both \cite{Cote:2018} \& \cite{Hausknecht:2019jericho} are specified in detail in this section.

\textbf{Partial Observability} 

The main challenge for agents solving Textual Games is the environment's partial observability; when observed information is only representative of a small part of the underlying truth. The difference between the two is often unknown and can require extensive exploration and failures for an agent to understand the game's connections between what is observed and how this relates to its actions.

A related additional challenge is that of causality which is when an agent moves away from a state to the next without completing pre-requisites of future states. For example, an agent is required to use a lantern necessary to light its way but may have to backtrack to previous states if this has not been obtained yet. An operation which becomes more complex with time and increases as the length of the agent's trajectory increases.

Handcrafted reward functions have been proved to work for easier settings, like CoinCollector \cite{Yuan:2018}, but real-world text games can require more nuanced approaches. Go-Explore has been used to find high-reward trajectories and discover under-explored states \cite{Madotto:2020,Ammanabrolu:2020grue} where more advanced states are given higher priority over states seen early on in the game by a weighted random exploration strategy.

\cite{Ammanabrolu:2020grue} have expanded on this with a modular policy approach aimed at noticing bottleneck states with a combination of a patience parameter and intrinsic motivation for new knowledge. The agent would learn a chain of policies and backtrack to previous states upon getting stuck.

Heuristic-based approaches have been used by \cite{Hausknecht:2019nail} to restrict navigational commands to only after all other interactive commands have been exhausted.

Leveraging past information has been proven to improve model performance as it allows to limit the partial observability aspect of the games. \cite{Ammanabrolu:2020graph} propose using a dynamically learned Knowledge Graph (KG) with a novel graph mask to only fill out templates with entities already in the learned KG.

\textbf{Large State Space} 

Whilst Textual Games, like all RL problems, require some form of exploration to find better solutions, some papers focus specifically on countering that natural overfitting of RL by actively encouraging exploration to unobserved states in new environments. For example, \cite{Yuan:2018} achieved this by setting a reward signal with a bonus for encountering a new state for the first time. This removes the agents capability for high-level contextual knowledge of the environment in favor of simply searching for unseen states. 

Subsequent work by \cite{Cote:2018} - Treasure Hunter - has expanded on this by increasing the complexity of the environment with additional obstacles such as locked doors that require colour matching keys requiring basic object affordance and generalisation ability.

In a similar vein to Treasure Hunter, where in the worst case agents have to traverse all states to achieve the objective, the \textit{location} and \textit{existence} settings of QAit \cite{Yuan:2019} require the same with addition of stating the location or existence of an object in the generated game. 

These solutions are also related to the challenge of \textit{Exploration vs Exploitation} that is commonly referenced in all RL literature \cite{Sutton:1998}.

\textbf{Large, Combinatorial and Sparse Action Spaces} 

Without any restrictions on length or semantics, RL agents aiming to solve games in this domain face the problem of an unbounded action space. Early works limited the action phrases to two words sentences for a verb-object, more recently combinatory action spaces are considered that include action phrases with multiple verbs and objects. A commonly used method for handling combinatory action spaces has been to limit the agent to picking a template $T$ and then filling in the blanks with entity extraction \cite{Hausknecht:2019jericho,Ammanabrolu:2020grue,Guo:2020}. 

Two other approaches have been used: (i) action elimination \cite{Zahavy:2018,Jain:2020}, and (ii) generative models \cite{Tao:2018,Yao:2020}. The first aims to use Deep Reinforcement Learning with an Action Elimination Network for approximation of the admissibility function: whether the action taken changes the underlying game state or not. The second has been used in limited scope with pointer softmax models generating commands over a fixed vocabulary and the recent textual observation. The CALM generative model, leveraging a fine-tuned GPT-2 for textual games, has proved to be competitive against models using valid action handicaps.


\textbf{Long-Term Credit Assignment} 

Assigning rewards to actions can be difficult in situations when the reward signals that are sparse. Specifically, positive rewards might only obtained at the successful completion of the game. However, environments where an agent is unlikely to reach the positive end game through random exploration, provide rewards for specific subtasks such as in \cite{Murugesan:2020env,Trischler:2019,Hausknecht:2019jericho}. The reward signal structured in this way also aligns with hierarchical approaches such as in \cite{Adolphs:2020}. Lastly, to overcome the challenges presented with reward-sparsity, various hand-crafted reward signals have been experimented with \cite{Yuan:2018,Ammanabrolu:2020grue}.


\textbf{Understanding Parser Feedback \& Language Acquisition} 

LIGHT \cite{Urbanek:2019}, a crowdsourced platform for the experimentation of grounded dialogue in fantasy settings that differs from the previous action-oriented environments by requiring dialogue with humans, embodied agents and the world itself as part of the quest completion. The authors design LIGHT to investigate how `a model can both speak and act grounded in perception of its environment and dialogue from other speakers'.

\cite{Ammanabrolu:2020light} extended this by providing a system that incorporates `1) large-scale language modelling based commonsense reasoning pre-training to imbue the agent with relevant priors and 2) a factorized action space of commands and dialogue'. Furthermore, evaluation can be performed against a dataset collected of held-out human demonstrations. 

\textbf{Commonsense Reasoning \& Affordance Extraction} 

As part of their semantic interpretation, Textual Games require some form of commonsense knowledge to be solved. For example, modeling the association between actions and associated objects (\textit{opening} doors instead of \textit{cutting} them, or the fact that \textit{taking} items allows to use them later on in the game). Various environments have been proposed for testing procedural knowledge in more distinct domains and to asses the agent's generalisation abilities.

For example, \cite{Trischler:2019} proposed the `First TextWorld Problems' competition with the intention of setting a challenge requiring more planning and memory than previous benchmarks. To achieve this, the competition featured `thousands of unique game instances generated using the TextWorld framework to share the same overarching theme - an agent is hungry in a house and has a goal of cooking a meal from gathered ingredients'. The agents therefore face a task that is more hierarchical in nature as cooking requires abstract instructions that entail a sequence of high-level actions on objects that are solved as sub-problems.

Furthermore, TW-Commonsense \cite{Murugesan:2020env} is explicitly built around agents leveraging prior commonsense knowledge for object-affordance and detection of out of place objects.

Two pre-training datasets have been proposed that form the evaluation goal for specialised modules of RL agents. The ClubFloyd dataset\footnote{\url{http://www.allthingsjacq.com/interactive_fiction.html\#clubfloyd}} provides human playthroughs of 590 different text based games, allowing to build priors and pre-train generative action generators. Likewise, the Jericho-QA \cite{Ammanabrolu:2020grue} dataset provides context at a specific timestep in various classical IF games supported by Jericho, and a list of questions, enabling pre-training of QA systems in the domain of Textual Games. They also used the dataset for fine-tuning a pre-trained LM for building a question-answering-based Knowledge Graph.

Lastly, ALFWorld \cite{Shridhar:2020} offers a new dimension by enabling the learning of a general policy in a textual environment and then test and enhance it in an embodied environment with a common latent structure. The general tasks also require commonsense reasoning for the agent to make connections between items and attributes, e.g. (sink, ``clean"), (lamp, ``light"). Likewise, the \textit{attribute} setting of QAit \cite{Yuan:2019} environment demands agents to understand attribute affordances (cuttable, edible, cookable) to find and alter (cut, eat, cook) objects in the environment.

\textbf{Knowledge Representation} 

It can be specified that at any given time step, the game's state can be represented as a graph that captures observation entities (player, objects, locations, etc) as vertices and the relationship between them as edges. As Text Games are partially observable, an agent can track its belief of the environment into a knowledge graph as it discovers it, eventually converging to an accurate representation of the entire game state \cite{Ammanabrolu:2018dqn}.

In contrast to methods which encode their entire KG into a single vector (as shown in \cite{Ammanabrolu:2020graph,Ammanabrolu:2020grue}) \cite{Xu:2020sha} suggest an intuitive approach of using multiple sub-graphs with different semantical meanings for multi-step reasoning. Previous approaches have relied on predefined rules and Stanford's Open Information Extraction \cite{Angeli:2015} for deriving information from observations for KG construction, \cite{Adhikari:2020} have instead built an agent, which is capable of building and updating it's belief graph without supervision.

Reframing the domain of Textual-Games, \cite{Guo:2020} consider observations as passages in a multi-passage RC task. They use an object-centric past-observation retrieval to enhance current state representations with relevant past information and then apply attention to draw focus on correlations for action-value prediction.

\section{Benchmark Environments and Agents}

Thus far, the majority of researchers have independently generated their own environments with TextWorld \cite{Cote:2018}. As the field moves towards more uniformity in evaluation, a clear overview of which environments are already generated, their design goals and the benchmark approach is needed.

Table \ref{tasks2} shows the publicly available environments and datasets. Much of the recent research has been published within the past two years (2018-2020). Jericho's Suite of Games (SoG) \cite{Hausknecht:2019jericho} is a collection of 52 games and therefore has its results averaged across all games included in evaluation.

We find that many of the environments focus on exploring increasingly complex environments to \textit{`collect'} an item of some kind ranging from a simple coin to household objects. This is due to the TextWorld generator well defined logic rules for such objects but can also be a limitation on the possible scope of the environment.

When evaluating an agent's performance, three types of evaluation settings are typically considered:

\begin{itemize}
    \item \textbf{Single Game} evaluate agents in the same game under the same conditions,
    \item \textbf{Joint} settings evaluate agents trained the same \textit{set} of games that typically share some similarities in the states seen,
    \item \textbf{Zero-Shot} settings evaluate agents on games completely unseen in training.
\end{itemize}

The difficulty settings of the environments are defined by the complexity of the challenges that the agents are required to overcome. In most cases, these have been limited to just a few levels. However, CookingWorld defines the challenge by a set of key variables that can be changed seperately or in combination and therefore does not offer clear discrete difficulty settings.

The max number of rooms \& objects depends heavily on the type of task. Coin Collecter for example has only 1 object to find as the complexity comes from traversing a number of rooms. Alternatively, CookingWorld has a limited number of rooms and instead focuses on a large number of objects to consider. Zork is naturally the most complex in this regards as an adaptation of a game designed for human players. Likewise, the complexity of the vocabulary depends heavily on the task but is clear to see that the environments limit the number of Action-Verbs for simplicity.
 


\begin{table}[H]
    \begin{center}
        \resizebox{\textwidth}{!}{%
            \begin{tabular}{l|l|c||c|c|c|c|c|c|c|c}
		        Name  & Task description & Eval   & GEN & \#Diffic.     & max $\#rooms$        & max $\#objects$ & $|AVS|$  & $\overline{len(o_t)}$            & $|V|$  & max $|quest|$    \\ 
                \midrule
                \midrule
                Zork I \cite{zork:1980} &  \textit{Collect} the \textbf{Twenty Treasures} of Zork & S      & N   & 1       & 110 & 251  & 237 & NS   & 697    & 396 \\
                Treasure Hunter \cite{Cote:2018} & \textit{Collect} \textbf{varying items} in a Home                    & ZS     & Y   & 30      & 20       & 2         & $\sim 4$  & NS & NS  & NS    \\
                Coin Collector  \cite{Yuan:2018} & \textit{Collect} \textbf{a coin} in a Home                       & S,J,ZS & Y   & 3       & 90      & 1         & 2    & 64 $\pm$ 9          & NS     & 30  \\
                FTWP/CookingWorld  \cite{Trischler:2019} & \textit{Select \& Combine} \textbf{varying items} in a Kitchen & J, ZS  & N   & Various & 12 & 56 & 18  & 97 $\pm$ 49       & 20,000     & 72       \\
                Jericho's SoG  \cite{Hausknecht:2019jericho} & A set of classical TGs  & S  & N   & 3 & NS         & $221_{avg}$           & NS  & $42,2_{avg}$     & $762_{avg}$   & $98_{avg}$  \\
                QAit  \cite{Yuan:2019} & Three QA type settings & ZS     & Y   & 1       & 12       & 27        & 17    & 93.1       & 1,647   & NS    \\
                TW-Home  \cite{Ammanabrolu:2018dqn} & \textit{Find} \textbf{varying objects} in a home & ZS & Y   & 2       & 20       & 40        & NS    & 94       & 819   & 10\\
                TW-Commonsense  \cite{Murugesan:2020env} & \textit{Collect} and \textit{Move}  \textbf{varying items} in a Home    & ZS  & Y   & 3          & 2        & 7 & $\sim 4$ & NS           & NS     & 17\\
                TW-Cook  \cite{Adhikari:2020} & \textit{Gather} and \textit{process} \textbf{cooking ingredients} & J, ZS & Y   & 5 & 9  & 34.1  & 28.4  & NS &  NS  & 3    \\
                \midrule
                TextWorld KG \cite{Zelinka:2019} & Constructing KGs from TG observations & - & N & 1 & N/A & N/A & N/A & 29.3 & NS & N/A  \\
                ClubFloyd  \cite{Yao:2020} & Human playthroughs of various TGs    & - & N   & 1   & N/A~ & N/A       & NS & NS & 39,670 & $360_{avg}$        \\
                Jericho-QA  \cite{Ammanabrolu:2020grue} & QA from context strings   & - & N   & 1       & N/A~           & N/A  & N/A    & $223.2_{avg}$  & NS   & N/A  
        \end{tabular}}
    \end{center}
    \caption{\small Environments for Textual-Games; Eval: (S)Single, (J) Joint, (ZS) Zero-Shot; GEN: engine support for generation of new games, \#Diffic.: number of difficulty settings, \#$rooms$ \& \#$objects$: number of rooms and objects per game, AVS: size of Action-Verb Space (NS=Not Specified), $\overline{len(o_t)}$: mean number of tokens in the observation $o_t$, $|V|$: Vocabulary size and, $|quest|$: length of optimal trajectory.}
    \label{tasks2}
\end{table}



\subsection{Agent Architectures}

\begin{figure}[h!]
    \centering
    \includegraphics[scale=0.45]{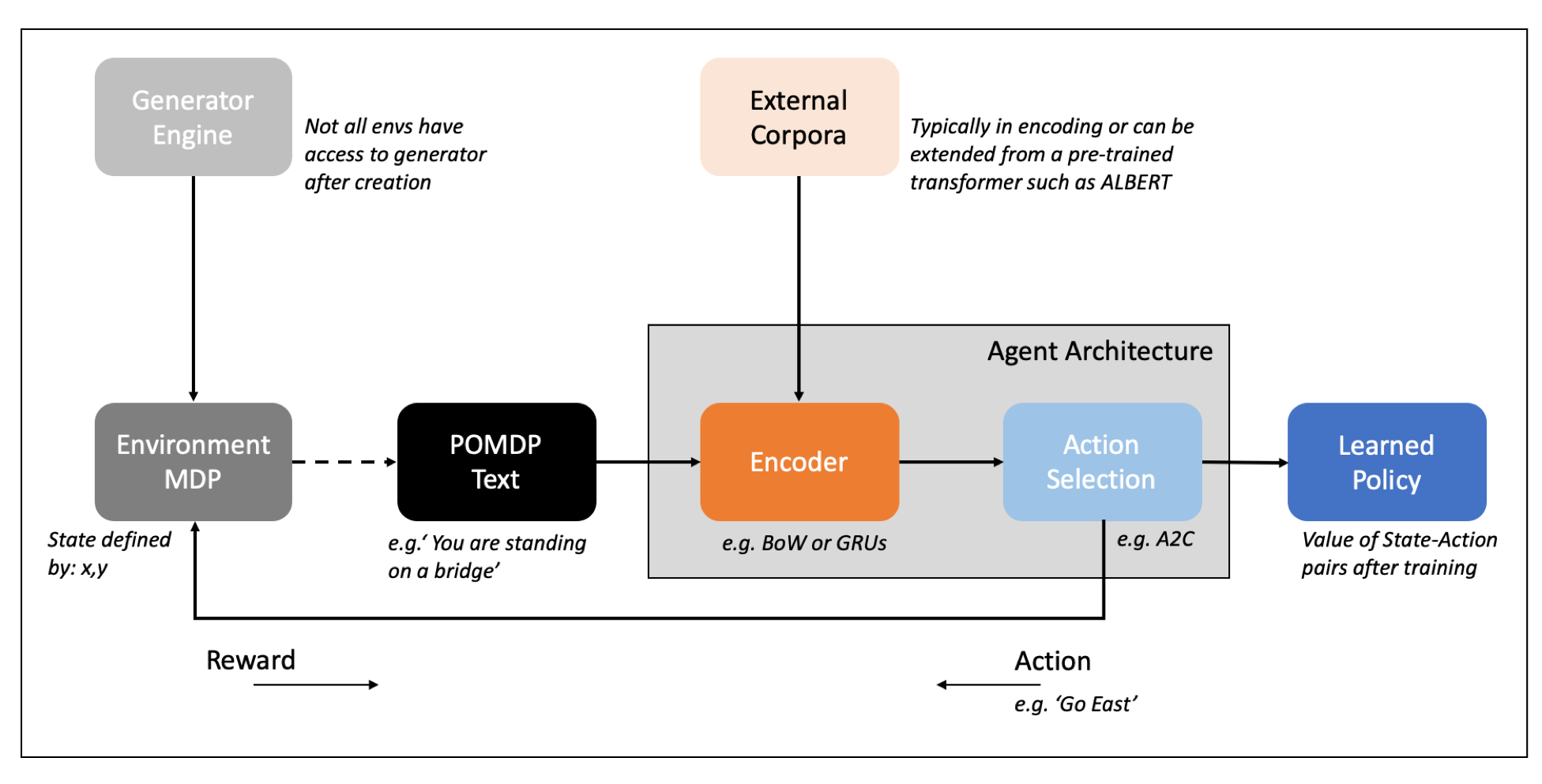}
    \caption{Overview of the Architecture Structure of Agents Applied to a Simple Text Game Example.}
    \label{fig:AgentArchi}
\end{figure}

A Deep-RL agent's architecture (see Figure \ref{fig:AgentArchi}) consists of two core components (i) state encoder and (ii) and action scorer \cite{Mnih:2013}. The first is used to encode game information such as observations, game feedback and KG representations into state approximations. The encoded information is then used by an action selection agent to estimate the value of actions in each state.

Table \ref{tab:Models} provides an overview of recent architectural trends for comparison. We find that the initial papers in 2015 used the standard approaches of LSTM or Bag of Words for the encoder and a Deep Q-Network (DQN) for the action selector. More recent developments have been experimenting with both parts towards improved results. Notably, a range of approaches for the encoding have been introduced with Gated Recurrent Unit's (GRUs) have become the most common in 2020. Whereas, there have been fewer variations in the choice of action selector where either Actor-Critic (A2C) or a DQN is typically used. Furthermore, the use of Knowledge Graphs and Pre-trained Tranformers is limited and many of the works that use these were published in 2020.

Most of the agents were applied to either TextWorld/Cookingworld or Jericho. We typically find that alternative environments align to consistent setups; either the authors create a new game that mimics simple real-world rules (similar to TextWorld) or they apply their methods to well know pre-existing games (similar to Jericho). Specifically, \cite{Narasimhan:2015} used a self generated two games themselves: `Home World' to mimic the environment of a typical house and `Fantasy World' that is more challenging and akin to a role-playing game. Alternatively, \cite{He:2015} used a deterministic text game `Saving John' and and a larger-scale stochastic text game `Machine of Death', both pre-existing from a public libary.

\begin{table}[H]
    \centering
    \resizebox{\textwidth}{!}{%
    \begin{tabular}{l||c|c|c|c||c|c||c}
         Name           & Encoder & Action Selector & KG  & PTF    & Pre-Training & AS & Tasks\\
         \midrule
         \midrule
         \cite{Ammanabrolu:2020grue}      & GRU & A2C   & DL    & ALBERT & J-QA & TB & Zork1\\
         \cite{Xu:2020sha}         & GRU & A2C   & DL    & none   & none & TB & JSoG\\
         \cite{Ammanabrolu:2020graph}  & GRU & A2C   & DL    & none   & ClubFloyd   & TB & JSoG\\
         \cite{Murugesan:2020model} & GRU & A2C & DL+CS & none & GloVe  & CB & TW-Commonsense\\
         \cite{Yao:2020}    & GRU & DRRN & none     & GPT-2  & ClubFloyd   & CB & JSoG\\
         \cite{Adolphs:2020} & Bi-GRU & A2C & none  & none   & TS, $\text{GloVe}_{100}$   & TB & CW\\
         \cite{Guo:2020}       & Bi-GRU & DQN & none & none   & $\text{GloVe}_{100}$ & TB & JSoG\\
         \cite{Xu:2020tf}   & TF  & DRRN  & none  & none   & none & TB & JSoG\\  
         \cite{Yin:2020snn}           & TF,LSTM & DSQN & none& none   & none & NS & CW, TH\\
         \cite{Adhikari:2020}           & R-GCN, TF & DDQN & DL & none & TS & CB & TW-Cook\\
         \cite{Zahavy:2018}         & CNN   & DQN & none & none & $\text{word2vec}_{300}$ & CB & Zork1\\
         \cite{Ammanabrolu:2018dqn} & LSTM & DQN & DL & none & TS, $\text{GloVe}_{100}$ & CB & TW-Home\\
         \cite{He:2015}           & BoW & DRRN   & none  & none & none   & CB & other\\
         \cite{Narasimhan:2015}       & LSTM & DQN  & none  & none   & none & PB & other\\
         \midrule
         \cite{Yin:2020bert}       & BERT & DQN  & DL  & BERT   & none & CB & FTWP, TH\\
         \cite{Madotto:2020} & LSTM & Seq2Seq & none & none & $\text{GloVe}_{100,300}$ & PB & CC, CW\\
    \end{tabular}}
    \caption{\small ENC - state/action encoder, KG - knowledge graph (DL: dynamically learned, CS: commonsense), PTF - pretrained Transformer, PreTr - pretraining (TS: task specific), AS - Action space (TB:template-based, PB: parser based, CB: choice-based)}
    \label{tab:Models}
\end{table}

\textbf{Encoders} used include both simplistic state encodings in the form of Bag of Words (BoW), but also recurrent modules like: Gated Recurrent Unit (GRU) \cite{Cho:2014}, Long Short-Term Memory (LSTM) \cite{Hochreiter:1997}, Transformer (TF) \cite{Vaswani:2017}, Relational Graph Convolutional Network (R-GCN) \cite{Schlichtkrull:2018}. Recently, Advantage Actor Critic (A2C) \cite{Mnih:2016} has gained popularity for the action selection method, with variants of the Deep Q-Network \cite{Mnih:2013}, such as the Deep Reinforcement Relevance Network  (DRRN) \cite{He:2015}, Double DQN (DDQN) \cite{Hasselt:2016}, Deep Siamese Q-Network (DSQN) \cite{Yin:2020snn}.







\textbf{Task Specific Pre-Training} entails heuristically establishing a setting in which a submodule learns priors before interacting with training data. For example, \cite{Adolphs:2020} pretrain on a collection of food items to improve generalizability to unseen objects and \cite{Adhikari:2020} pretrain a KG constructor on trajectories of similar games the agent is trained on. \cite{Chaudhury:2020} showed that training an agent on pruned observation space, where the semantically least relevant tokens are removed in each episode to improve generalizability to unseen domains whilst also improving the sample efficieny due to requiring less training games. Furthermore, \cite{Jain:2020} propose learning different action-value functions for all possible scores in a game, thus effectively learning separate value functions for each subtask of a whole.    \

Lastly, the \textbf{action-space} typically varies between template or choice based depending on the type of task. Only two papers have considered a parser based approach: \cite{Narasimhan:2015} and \cite{Madotto:2020}.

\subsection{Benchmark Results}

The following section summarises some of the results published thus far as a means to review the performance of the baselines as they are presented by the original papers to be used for future comparisons.

\textbf{Treasure Hunter} was introduced by \cite{Cote:2018} as part of the TextWorld formalisation and inspired by a classic problem to navigate a maze to find a specific object. The agent and objects are placed in a randomly generated map. A coloured object near the agent's start location provides an indicator of which object to obtain provided in the welcome message. A straight forward reward signal is defined as positive for obtaining the correct object and negative for an incorrect object with a limited number of turns available. 

Increasing difficulties are defined by the number of rooms, quest length and number of locked doors and containers. Levels 1 to 10 have only 5 rooms, no doors and an increasing quest length from 1 to 5. Levels 11 to 20 have 10 rooms, include doors and containers that may need to be opened and quest length increasing from 2 to 10. Lastly, levels 21 to 30 have 20 rooms, locked doors and containers that may need to be unlocked and opened and quest length increasing from 3 to 20. 

For the evaluation, two state-of-the-art agents (BYU \cite{Fulda:2017} and Golovin \cite{Kostka:2017}) were compared to a choice-based random agent in a \textit{zero-shot} evaluation setting. For completeness, each was applied to 100 generated games at varying difficulty levels up to a maximum of 1,000 steps. The results of this are shown in table \ref{tab:TreasureHunterResults} where we note that the random agent performs best but this is due to the choice-based method removing the complexity of the compositional properties of language. The authors noted that may not be directly comparable to the other agents but provides an indication of the difficulty of the tasks.

\begin{table}[H]
    \centering
    \resizebox{\textwidth}{!}{%
    \begin{tabular}{l||c c|c c|c c}
         &\textbf{Random} & & \textbf{BYU} & & \textbf{Golovin} & \\
         Difficulty & Avg. Score & Avg. Steps & Avg. Score & Avg. Steps & Avg. Score & Avg. Steps\\
         \midrule
         \midrule
         level 1  & 0.35  & 9.85   & 0.75 & 85.18   & 0.78  & 18.16\\
         level 5  & -0.16 & 19.43  & -0.33 & 988.72 & -0.35 & 135.67\\
         level 10 & -0.14 & 20.74  & -0.04 & 1000   & -0.05 & 609.16\\
         level 11 & 0.30  & 43.75  & 0.02 & 992.10  & 0.04  & 830.45\\
         level 15 & 0.27  & 63.78  & 0.01 & 998     & 0.03  & 874.32\\
         level 20 & 0.21  & 74.80  & 0.02 & 962.27  & 0.04  & 907.67\\
         level 21 & 0.39  & 91.15  & 0.04 & 952.78  & 0.09  & 928.83\\
         level 25 & 0.26  & 101.67 & 0.00 & 974.14  & 0.04  & 931.57\\
         level 30 & 0.26  & 108.38 & 0.04 & 927.37  & 0.74  & 918.88\\
    \end{tabular}}
    \caption{Results of agents applied to Treasure Hunter's one-life tasks.}
    \label{tab:TreasureHunterResults}
\end{table}

\textbf{CookingWorld} is the second well-established environment developed using TextWorld by \cite{Trischler:2019}, is used in \textit{joint} and \textit{zero-shot} settings, which both enable testing for generalization. The former entails training and evaluating the agent on the same set of games, while the latter uses an unseen test set upon evaluation. The fixed dataset provides $4,400$ training, $222$ validation and $514$ test games with $222$ different types of games in various game difficulty settings and the current best results are stated in Table \ref{tab:CookingWorld}. 
However there are some variances in the split of training and test games, making it challenging to compare the results. Most models were trained using the games split of 4,400/514 specified before, but there are variances as some split it as 3,960/440 and some do not state the split at all.
Table \ref{tab:CookingWorld} shows the results of this comparison with a percentage score relative to the maximum reward.


\begin{table}[H]
    \centering
    \resizebox{\textwidth}{!}{%
    \begin{tabular}{l||c|c|c|c||c|c}
         & DSQN & LeDeepChef & Go-Explore Seq2Seq & BERT-NLU-SE & LSTM-DQN & DRRN\\
         \midrule
         \midrule
         Single & - & - & 88.1\% & - & 52.1\% & 80.9\%\\
         Joint  & - & - & 56.2\% &  - & 3.1\% & 22.9\%\\
         Zero-Shot & 58\% & 69.3\% & 51\% & 77\% & 2\% & 22.2\%\\
         Treasure Hunter (OoD) & 42\% & - & - & 57\% & - & - \\
         \% won games (Zero-Shot) & - & - & 46.6\% & 71\% & 3.2\% & 8.3\% \\
         avg \#steps (Zero-Shot) & - / 100 & 43.9 / 100 & 24.3 / 50 & - / 100 & 48.5 / 50 & 38.8 / 50 \\
         \midrule
         \#training steps & $10^{7}$ & 13,200 & NS & $10^{7}_{Tchr}$, $5 * 10^{5}_{stdn}$ & NS & NS \\
         Admissable Actions & NS & N & N & Y & Y & Y\\
         Curriculum Learning & N & N & Y & Y & N & N\\
         Imitation Learning & N & N & Y & Y & N & N\\
         Training time & NS & NS & NS & Tchr: 35 days, Stdn: 5 days & NS & NS\\
    \end{tabular}}
    \caption{Results on the public CookingWorld dataset; OoD - out of domain}
    \label{tab:CookingWorld}
\end{table}



\textbf{Jericho's Suite of Games} was defined by \cite{Hausknecht:2019jericho} and has been used for testing agent's in a \textit{single game} setting. As the suite provides 57 games, of which a variation of 32 are used due to excessive complexity, then the agents are initialized, trained and evaluated on each of them separately. Agent's are assessed over the \textit{final score}, which is by convention averaged over the last 100 episodes (which in turn is usually averaged over 5 agents with different initialization seeds). The current state of the art results on the Jericho gameset can be seen in Table \ref{tab:JerichoGames} where numeric results are shown relative to the Maximum Reward with the aggregated average percentage shown in the final row\footnote{NAIL is a rule-based agent and is noted to emphasize the capabilities of learning based models}. 

\begin{table}[H]
    \centering
    \resizebox{\textwidth}{!}{%
    \begin{tabular}{l||c||c|c|c|c|c|c|c||c||c|c}
         & MaxR & CALM-DRRN & SHA-KG    & Trans-v-DRRN  & MPRC-DQN   & KG-A2C   & TDQN & DRRN & NAIL  & $|T|$ & $|V|$ \\
         \midrule
         \midrule
         \textcolor{blue}{905}      & 1       & \textbf{0}         & -   & \textbf{0}  & \textbf{0}  & \textbf{0}    & \textbf{0} & \textbf{0} & \textbf{0} & 82    & 296\\
         \textcolor{blue}{acorncourt} & 30 & 0         & 1.6 & \textbf{10} & \textbf{10}        & 0.3 & 1.6 & \textbf{10} & 0 & 151   & 343\\
         \textcolor{Dandelion}{advent} & 350    & 36        & - & - & \textbf{63.9} & 36 & 36 & 20.6 & 36 & 189 & 786\\
         \textcolor{blue}{advland} & 100 & 0 & - & 25.6 & \textbf{42.2} & 0 & 0 & 20.6 & 0 & 156 & 398 \\
         \textcolor{blue}{affilicted} & 75& - & - & 2.0 & \textbf{8.0} & - & 1.4 & 2.6 & 0 & 146 & 762\\
         \textcolor{red}{anchor} & 100 & \textbf{0} &- & - & \textbf{0} & \textbf{0} & \textbf{0} & \textbf{0} & \textbf{0} & 260 & 2257\\
         \textcolor{blue}{awaken} & 50 & \textbf{0} & - & - & \textbf{0} & \textbf{0} & \textbf{0} & \textbf{0} & \textbf{0} & 159 & 505\\
         \textcolor{Dandelion}{balances} & 51 & 9.1 & \textbf{10.0} & - & \textbf{10} & \textbf{10} & 4.8 & \textbf{10} & 10 & 156 & 452\\
         \textcolor{Dandelion}{deephome} & 300 & 1 & - & - & 1 & 1 & 1 & 1 & \textbf{13.3} & 173 & 760\\
         \textcolor{blue}{detective} & 360 & 289.7 & 208.0 & 288.8 & \textbf{317.7} & 207.9 & 169 & 197.8 & 136.9 & 197 & 344\\
         \textcolor{blue}{dragon} & 25 & 0.1 & 0.2 & - & 0.04 & 0 & -5.3 & -3.5 & \textbf{0.6} & 177 & 1049\\
         \textcolor{red}{enchanter} & 400 & 19.1 & \textbf{20} & \textbf{20} & \textbf{20} & 1.1 & 8.6 & \textbf{20} & 0 & 290 & 722\\
         \textcolor{Dandelion}{gold} & 100 & - & - & - & 0 & - & \textbf{4.1} & 0 & 3 & 200 & 728\\
         \textcolor{blue}{inhumane} & 90 &  \textbf{25.7} & 5.4 & - & 0 & 3 & 0.7 & 0 & 0.6 & 141 & 409\\
         \textcolor{Dandelion}{jewel} & 90 &  0.3 & 1.8 & - & \textbf{4.46} & 1.8 & 0 & 1.6 & 1.6 & 161 & 657\\
         \textcolor{Dandelion}{karn} & 170 & 2.3 & - & - & \textbf{10} & 0 & 0.7 & 2.1 & 1.2 & 178 & 615\\
         \textcolor{blue}{library}  & 30 &  9.0 & 15.8 & 17 & \textbf{17.7} & 14.3 & 6.3 & 17 & 0.9 & 173 & 510\\
         \textcolor{Dandelion}{ludicorp} & 150 & 10.1 & 17.8 & 16 & \textbf{ 19.7} & 17.8 & 6 & 13.8 & 8.4 & 187 & 503\\
         \textcolor{blue}{moonlit}  & 1 & \textbf{0} & - & - & \textbf{0} & \textbf{0} & \textbf{0} & \textbf{0} & \textbf{0} & 166 &  669\\
         \textcolor{blue}{omniquest}& 50 & 6.9 & - & - & 10 & 3 & \textbf{16.8} & 5  & 5.6 & 207 & 460\\
         \textcolor{blue}{pentari} & 70 & 0 & \textbf{51.3} & 34.5 & 44.4 & 50.7 & 17.4 & 27.2 & 0 & 155 & 472\\
         \textcolor{blue}{reverb} & 50 & - & 10.6 &\textbf{ 10.7} & 2.0 & - & 0.3 & 8.2 & 0 & 183 & 526\\
         \textcolor{blue}{snacktime}  & 50 & \textbf{19.4} & - & - & 0 & 0 & 9.7 & 0 & 0 & 201 & 468\\
         \textcolor{red}{sorcerer}  & 400 & 6.2 & 29.4 & - & \textbf{38.6} & 5.8 & 5 & 20.8 & 5 & 288 & 1013\\
         \textcolor{red}{spellbrkr}  & 600 & \textbf{40} & \textbf{40} & \textbf{40} & 25 & 21.3 & 18.7 & 37.8 & \textbf{40} & 333 & 844\\
         \textcolor{red}{spirit}  & 250 & 1.4 & \textbf{3.8} & - & \textbf{3.8} & 1.3 & 0.6 & 0.8 & 1 & 169 & 1112\\
         \textcolor{blue}{temple}  & 35 & 0 & 7.9 & 7.9 & \textbf{8.0} & 7.6 & 7.9 & 7.4 & 7.3 & 175 & 622\\
         \textcolor{red}{tryst205} & 350 & - & 6.9 & 9.6 & \textbf{10} & - & 0 & 9.6  & 2 & 197 & 871\\
         \textcolor{Dandelion}{yomomma} & 35 - & - & - & - & \textbf{1} & - & 0 & 0.4 & 0 & 141 & 619 \\
         \textcolor{Dandelion}{zenon}  & 20 & 0 & \textbf{3.9} & - & 0 & \textbf{3.9} & 0 & 0 & 0 & 149 & 401\\
         \textcolor{Dandelion}{zork1} & 350 & 30.4 & 34.5 & 36.4 & \textbf{38.3} & 34 & 9.9 & 32.6  & 10.3 & 237 & 697\\
         \textcolor{Dandelion}{zork3} & 7 & 0.5 & 0.7 & 0.19 & \textbf{3.63} & 0.1 & 0 & 0.5 & 1.8 & 214 & 564\\
         \textcolor{blue}{ztuu} & 100 & 3.7 & 25.2 & 4.8 & \textbf{85.4} & 9.2 & 4.9 & 21.6  & 0 & 186 & 607\\
         \midrule
         Winning \% / count  & & 21.2\%/ 7  & 18.2\%/ 6 &  15.2\%/ 5 &  69.7\%/ 23  &  18.2\%/ 6 &  18.2\%/ 6 &  21.2\%/ 7 & 21.2\%/ 7  \\
         Avg. norm / \# games & &  9.4\% / 28 & 19.6\% / 20 & 22.3\% / 15 & 17\% / 33 & 10.8\% / 28  & 6\% / 33 & 10.7\% / 32  & 4.8\% / 33\\
         \midrule
         Handicaps & & $\{1\}$ & $\{1, 2, 4\}$ & NS & $\{1, 2, 4\}$ & $\{1, 2, 4\}$ & $\{1, 2, 4\}$ & $\{1, 4\}$ & N/A \\
         Train steps & & $10^{6}$ & $10^{6}$ & $10^{5}$ & $10^{5}$ & $1.6 \times 10^{6}$ & $10^{6}$ & $1.6 \times 10^{6}$ & N/A \\
         Train time & & NS & NS & NS & 8h-30h per game & NS & NS & NS & N/A\\
    \end{tabular}}
    \caption{\small \textcolor{blue}{Blue}: likely to be solved in near future. \textcolor{Dandelion}{Orange}: progress is likely, significant scientific progress necessary to solve. \textcolor{Red}{Red}: very difficult even for humans, unthinkable for current RL. MaxR - Maximum possible reward ;$|T|$ - number of templates; $|V|$ - size of vocabulary set} 
    \label{tab:JerichoGames}
\end{table}


\section{Conclusion and Future Work}
  
Recent research developments for interactive fiction games have generated a set of environments and tested their architectures for generalization and overfitting capabilities. With the comparisons in this work and a uniform set of environments and baselines, new architectures can be systematically contrasted. Currently, most methodologies focus on either an agent's ability to learn efficiently or generalise knowledge. A continued development in this direction is the primary motive for much of the recent research and could lead to solutions that work sufficiently well for the primary real-world challenge of limited and expensive data.

The most recent advances the agent's performance to reach a goal given the linguistic challenge have come from utilising pre-trained transformer models. This has been supported by work published in the NLP community with methods such as BERT \cite{Yin:2020bert}, GPT-2 \cite{Yao:2020} and ALBERT\cite{Ammanabrolu:2020grue} and continued developments from this community will support the advancements of future agents architectures. 

Similarly, the use of dynamically learned knowledge graphs have become common in recent works and further developments from the NLP community (such as \cite{Das:2019}) could enable the use of pre-training knowledge graphs on readily available text corpora before training on the environment itself.  

However, there are no methods that currently consider interpretability with language as a primary goal. This is not a challenge unique to Text Games or Reinforcement Learning as calls for research have been made in similar machine learning domains \cite{Chu:2020}. A notable methodology that is worth considering is the Local Interpretable Model-Agnostic Explanations (LIME) introduced by \cite{lime:2016}.

Furthermore, as noted by \cite{luketina:2019}, grounding language into the environments is still an open problem that has yet to be addressed fully. \cite{Urbanek:2019} introduces a Text Game environment specifically for grounding dialogue as a first step but this is still an open problem. 

Lastly, from the contributions analysed in this survey only 5 papers report the amount of time and resources their methods needed for training. Continuing the trend of publishing these specifications is essential in making results more reproducible and applicable in applied settings. Reducing the amount of resources and training time required as a primary motive allows for the domain to be practical as a solution and also accessible given that not all problems allow for unlimited training samples on high performance machines.

\bibliographystyle{unsrtnat}

\begin{thebibliography}{48}
\providecommand{\natexlab}[1]{#1}
\providecommand{\url}[1]{\texttt{#1}}
\expandafter\ifx\csname urlstyle\endcsname\relax
  \providecommand{\doi}[1]{doi: #1}\else
  \providecommand{\doi}{doi: \begingroup \urlstyle{rm}\Url}\fi


\bibitem[Adhikari et~al.(2020)Adhikari, Yuan, C{\^o}t{\'e}, Zelinka, Rondeau,
  Laroche, Poupart, Tang, Trischler, and Hamilton]{Adhikari:2020}
Ashutosh Adhikari, Xingdi Yuan, Marc-Alexandre C{\^o}t{\'e}, Mikul{\'a}{\v{s}}
  Zelinka, Marc-Antoine Rondeau, Romain Laroche, Pascal Poupart, Jian Tang,
  Adam Trischler, and Will Hamilton.
\newblock Learning dynamic belief graphs to generalize on text-based games.
\newblock \emph{Advances in Neural Information Processing Systems}, 33, 2020.

\bibitem[Adolphs and Hofmann(2020)]{Adolphs:2020}
Leonard Adolphs and T.~Hofmann.
\newblock Ledeepchef: Deep reinforcement learning agent for families of
  text-based games.
\newblock In \emph{AAAI}, 2020.


\bibitem[Ammanabrolu and Riedl(2019)]{Ammanabrolu:2018dqn}
Prithviraj Ammanabrolu and Mark~O Riedl.
\newblock Playing text-adventure games with graph-based deep reinforcement
  learning.
\newblock \emph{Proceedings of the 2019 Conference of the North American
  Chapter of the Association for Computational Linguistics: Human Language
  Technologies, Volume 1 (Long and Short Papers)}, pages 3557–--3565, 2019.


\bibitem[Ammanabrolu et~al.(2020{\natexlab{a}})Ammanabrolu, Tien, Hausknecht,
  and Riedl]{Ammanabrolu:2020grue}
Prithviraj Ammanabrolu, Ethan Tien, Matthew Hausknecht, and Mark~O. Riedl.
\newblock How to avoid being eaten by a grue: Structured exploration strategies
  for textual worlds.
\newblock volume abs/2006.07409, 2020{\natexlab{a}}.
\newblock URL \url{http://arxiv.org/abs/2006.07409}.

\bibitem[Ammanabrolu and Hausknecht(2020)]{Ammanabrolu:2020graph}
Prithviraj Ammanabrolu and Matthew~J. Hausknecht.
\newblock Graph constrained reinforcement learning for natural language action
  spaces.
\newblock \emph{International Conference on Learning Representations}, 2020.


\bibitem[Ammanabrolu et~al.(2020{\natexlab{b}})Ammanabrolu, Urbanek, Li, Szlam,
  Rocktaschel, and Weston]{Ammanabrolu:2020light}
Prithviraj Ammanabrolu, Jack Urbanek, Margaret Li, Arthur Szlam, Tim
  Rocktaschel, and J.~Weston.
\newblock How to motivate your dragon: Teaching goal-driven agents to speak and
  act in fantasy worlds.
\newblock \emph{ArXiv}, abs/2010.00685, 2020{\natexlab{b}}.


\bibitem[Anderson et~al.(1980)Anderson, Blank, Daniels, and Lebling]{zork:1980}
Tim Anderson, Marc Blank, Bruce Daniels, and Dave Lebling.
\newblock Zork: The great underground empire – part i, 1980.


\bibitem[Angeli et~al.(2015)Angeli, Johnson~Premkumar, and
  Manning]{Angeli:2015}
Gabor Angeli, Melvin~Jose Johnson~Premkumar, and Christopher~D. Manning.
\newblock Leveraging linguistic structure for open domain information
  extraction.
\newblock In \emph{Proceedings of the 53rd Annual Meeting of the Association
  for Computational Linguistics and the 7th International Joint Conference on
  Natural Language Processing (Volume 1: Long Papers)}, pages 344--354,
  Beijing, China, July 2015. Association for Computational Linguistics.
\newblock \doi{10.3115/v1/P15-1034}.
\newblock URL \url{https://www.aclweb.org/anthology/P15-1034}.

\bibitem[Chaudhury et~al.(2020)Chaudhury, Kimura, Talamadupula, Tatsubori,
  Munawar, and Tachibana]{Chaudhury:2020}
Subhajit Chaudhury, Daiki Kimura, Kartik Talamadupula, Michiaki Tatsubori, Asim
  Munawar, and Ryuki Tachibana.
\newblock Bootstrapped q-learning with context relevant observation pruning to
  generalize in text-based games.
\newblock 2020.


\bibitem[Cho et~al.(2014)Cho, Merrienboer, Çaglar G{\"u}lçehre, Bahdanau,
  Bougares, Schwenk, and Bengio]{Cho:2014}
Kyunghyun Cho, B.~V. Merrienboer, Çaglar G{\"u}lçehre, Dzmitry Bahdanau,
  Fethi Bougares, Holger Schwenk, and Yoshua Bengio.
\newblock Learning phrase representations using rnn encoder-decoder for
  statistical machine translation.
\newblock \emph{Empirical Methods in Natural Language Processing},
  abs/1406.1078, 2014.


\bibitem[Chu et~al.(2020)Chu, Gillani, and Priscilla~Makini]{Chu:2020}
Eric Chu, Nabeel Gillani, and Sneha Priscilla~Makini.
\newblock Games for fairness and interpretability.
\newblock In \emph{Companion Proceedings of the Web Conference 2020}, WWW '20,
  page 520–524, New York, NY, USA, 2020. Association for Computing Machinery.
\newblock ISBN 9781450370240.
\newblock \doi{10.1145/3366424.3384374}.
\newblock URL \url{https://doi.org/10.1145/3366424.3384374}.


\bibitem[C{\^o}t{\'e} et~al.(2018)C{\^o}t{\'e}, K{\'a}d{\'a}r, Yuan, Kybartas,
  Barnes, Fine, Moore, Hausknecht, Asri, Adada, Tay, and Trischler]{Cote:2018}
Marc-Alexandre C{\^o}t{\'e}, {\'A}kos K{\'a}d{\'a}r, Xingdi Yuan, Ben Kybartas,
  Tavian Barnes, Emery Fine, J.~Moore, Matthew~J. Hausknecht, Layla~El Asri,
  Mahmoud Adada, Wendy Tay, and Adam Trischler.
\newblock Textworld: A learning environment for text-based games.
\newblock \emph{Workshop on Computer Games}, pages 41--75, 2018.


\bibitem[Das et~al.(2019)Das, Munkhdalai, Yuan, Trischler, and
  McCallum]{Das:2019}
Rajarshi Das, Tsendsuren Munkhdalai, Xingdi Yuan, Adam Trischler, and Andrew
  McCallum.
\newblock Building dynamic knowledge graphs from text using machine reading
  comprehension.
\newblock \emph{ICLR}, 2019.


\bibitem[Dulac-Arnold et~al.(2019)Dulac-Arnold, Mankowitz, and
  Hester]{dulac:2019}
Gabriel Dulac-Arnold, Daniel Mankowitz, and Todd Hester.
\newblock Challenges of real-world reinforcement learning.
\newblock \emph{Proceedings of the 36 th International Conference on Machine
  Learning}, 2019.


\bibitem[Fulda et~al.(2017)Fulda, Ricks, Murdoch, and Wingate]{Fulda:2017}
Nancy Fulda, Daniel Ricks, Ben Murdoch, and David Wingate.
\newblock What can you do with a rock? affordance extraction via word
  embeddings.
\newblock \emph{arXiv preprint arXiv: 1703.03429}, 2017.


\bibitem[Guo et~al.(2020)Guo, Yu, Gao, Gan, Campbell, and Chang]{Guo:2020}
Xiaoxiao Guo, M.~Yu, Yupeng Gao, Chuang Gan, Murray Campbell, and S.~Chang.
\newblock Interactive fiction game playing as multi-paragraph reading
  comprehension with reinforcement learning.
\newblock \emph{EMNLP}, 2020.

\bibitem[Hasselt et~al.(2016)Hasselt, Guez, and Silver]{Hasselt:2016}
H.~V. Hasselt, A.~Guez, and D.~Silver.
\newblock Deep reinforcement learning with double q-learning.
\newblock \emph{AAAI}, 2016.


\bibitem[Hausknecht et~al.(2019{\natexlab{a}})Hausknecht, Ammanabrolu,
  C\^ot\'{e}, and Yuan]{Hausknecht:2019jericho}
Matthew Hausknecht, Prithviraj Ammanabrolu, Marc-Alexandre C\^ot\'{e}, and
  Xingdi Yuan.
\newblock Interactive fiction games: A colossal adventure.
\newblock \emph{CoRR}, abs/1909.05398, 2019{\natexlab{a}}.
\newblock URL \url{http://arxiv.org/abs/1909.05398}.

\bibitem[Hausknecht et~al.(2019{\natexlab{b}})Hausknecht, Loynd, Yang,
  Swaminathan, and Williams]{Hausknecht:2019nail}
Matthew Hausknecht, Ricky Loynd, Greg Yang, Adith Swaminathan, and Jason~D
  Williams.
\newblock Nail: A general interactive fiction agent.
\newblock \emph{arXiv preprint arXiv:1902.04259}, 2019{\natexlab{b}}.


\bibitem[He et~al.(2016)He, Chen, He, Gao, Li, Deng, and Ostendorf]{He:2015}
Ji~He, Jianshu Chen, Xiaodong He, Jianfeng Gao, Lihong Li, Li~Deng, and Mari
  Ostendorf.
\newblock Deep reinforcement learning with a natural language action space.
\newblock \emph{Association for Computational Linguistics (ACL)}, 2016.


\bibitem[Hochreiter and Schmidhuber(1997)]{Hochreiter:1997}
Sepp Hochreiter and J{\"u}rgen Schmidhuber.
\newblock Long short-term memory.
\newblock \emph{Neural computation}, 9\penalty0 (8):\penalty0 1735--1780, 1997.


\bibitem[Jain et~al.(2020)Jain, Fedus, Larochelle, Precup, and
  Bellemare]{Jain:2020}
Vishal Jain, William Fedus, Hugo Larochelle, Doina Precup, and Marc~G
  Bellemare.
\newblock Algorithmic improvements for deep reinforcement learning applied to
  interactive fiction.
\newblock In \emph{AAAI}, pages 4328--4336, 2020.


\bibitem[Kostka et~al.(2017)Kostka, Kwiecieli, Kowalski, and
  Rychlikowski]{Kostka:2017}
Bartosz Kostka, Jaroslaw Kwiecieli, Jakub Kowalski, and Pawel Rychlikowski.
\newblock Text-based adventures of the golovin ai agent.
\newblock \emph{Computational Intelligence and Games (CIG)}, pages 181--188,
  2017.

\bibitem[Luketina et~al.(2019)Luketina, Nardelli, Farquhar, Foerster, Andreas,
  Grefenstette, Whiteson, and Rockt{\"a}schel]{luketina:2019}
Jelena Luketina, Nantas Nardelli, Gregory Farquhar, Jakob Foerster, Jacob
  Andreas, Edward Grefenstette, Shimon Whiteson, and Tim Rockt{\"a}schel.
\newblock A survey of reinforcement learning informed by natural language.
\newblock \emph{arXiv e-prints}, 2019.

\bibitem[Madotto et~al.(2020)Madotto, Namazifar, Huizinga, Molino, Ecoffet,
  Zheng, Papangelis, Yu, Khatri, and Tur]{Madotto:2020}
Andrea Madotto, Mahdi Namazifar, Joost Huizinga, Piero Molino, Adrien Ecoffet,
  Huaixiu Zheng, Alexandros Papangelis, Dian Yu, Chandra Khatri, and Gokhan
  Tur.
\newblock Exploration based language learning for text-based games.
\newblock \emph{IJCAI}, 2020.

\bibitem[Mnih et~al.(2013)Mnih, Kavukcuoglu, Silver, Graves, Antonoglou,
  Wierstra, and Riedmiller]{Mnih:2013}
Volodymyr Mnih, Koray Kavukcuoglu, David Silver, Alex Graves, Ioannis
  Antonoglou, Daan Wierstra, and Martin~A. Riedmiller.
\newblock Playing atari with deep reinforcement learning.
\newblock \emph{CoRR}, abs/1312.5602, 2013.
\newblock URL \url{http://arxiv.org/abs/1312.5602}.

\bibitem[Mnih et~al.(2016)Mnih, Badia, Mirza, Graves, Lillicrap, Harley,
  Silver, and Kavukcuoglu]{Mnih:2016}
Volodymyr Mnih, Adria~Puigdomenech Badia, Mehdi Mirza, Alex Graves, Timothy
  Lillicrap, Tim Harley, David Silver, and Koray Kavukcuoglu.
\newblock Asynchronous methods for deep reinforcement learning.
\newblock pages 1928--1937, 2016.

\bibitem[Murugesan et~al.(2020{\natexlab{a}})Murugesan, Atzeni, Kapanipathi,
  Shukla, Kumaravel, Tesauro, Talamadupula, Sachan, and
  Campbell]{Murugesan:2020env}
Keerthiram Murugesan, Mattia Atzeni, Pavan Kapanipathi, Pushkar Shukla, Sadhana
  Kumaravel, Gerald Tesauro, Kartik Talamadupula, Mrinmaya Sachan, and Murray
  Campbell.
\newblock Text-based rl agents with commonsense knowledge: New challenges,
  environments and baselines.
\newblock \emph{CoRR}, abs/2010.03790, 2020{\natexlab{a}}.

\bibitem[Murugesan et~al.(2020{\natexlab{b}})Murugesan, Atzeni, Shukla, Sachan,
  Kapanipathi, and Talamadupula]{Murugesan:2020model}
Keerthiram Murugesan, Mattia Atzeni, Pushkar Shukla, Mrinmaya Sachan, Pavan
  Kapanipathi, and Kartik Talamadupula.
\newblock Enhancing text-based reinforcement learning agents with commonsense
  knowledge.
\newblock \emph{arXiv preprint arXiv:2005.00811}, 2020{\natexlab{b}}.

\bibitem[Narasimhan et~al.(2015)Narasimhan, Kulkarni, and
  Barzilay]{Narasimhan:2015}
Karthik Narasimhan, Tejas Kulkarni, and Regina Barzilay.
\newblock Language understanding for text-based games using deep reinforcement
  learning.
\newblock \emph{EMNLP}, 2015.


\bibitem[Ribeiro et~al.(2016)Ribeiro, Singh, and Guestrin]{lime:2016}
Marco~Tulio Ribeiro, Sameer Singh, and Carlos Guestrin.
\newblock "why should {I} trust you?": Explaining the predictions of any
  classifier.
\newblock In \emph{Proceedings of the 22nd {ACM} {SIGKDD} International
  Conference on Knowledge Discovery and Data Mining, San Francisco, CA, USA,
  August 13-17, 2016}, pages 1135--1144, 2016.

\bibitem[Schlichtkrull et~al.(2018)Schlichtkrull, Kipf, Bloem, Van Den~Berg,
  Titov, and Welling]{Schlichtkrull:2018}
Michael Schlichtkrull, Thomas~N Kipf, Peter Bloem, Rianne Van Den~Berg, Ivan
  Titov, and Max Welling.
\newblock Modeling relational data with graph convolutional networks.
\newblock pages 593--607, 2018.

\bibitem[Shridhar et~al.(2020)Shridhar, Yuan, C{\^o}t{\'e}, Bisk, Trischler,
  and Hausknecht]{Shridhar:2020}
Mohit Shridhar, Xingdi Yuan, Marc-Alexandre C{\^o}t{\'e}, Yonatan Bisk, Adam
  Trischler, and Matthew~J. Hausknecht.
\newblock Alfworld: Aligning text and embodied environments for interactive
  learning.
\newblock \emph{ArXiv}, abs/2010.03768, 2020.


\bibitem[Silver et~al.(2016)Silver, Huang, Maddison, Guez, Sifre, van~den
  Driessche, Schrittwieser, Antonoglou, Panneershelvam, Lanctot, Dieleman,
  Grewe, Nham, Kalchbrenner, Sutskever, Lillicrap, Leach, Kavukcuoglu, Graepel,
  and Hassabis]{Silver:2016}
D.~Silver, Aja Huang, Chris~J. Maddison, A.~Guez, L.~Sifre, George van~den
  Driessche, Julian Schrittwieser, Ioannis Antonoglou, Vedavyas Panneershelvam,
  Marc Lanctot, S.~Dieleman, Dominik Grewe, John Nham, Nal Kalchbrenner, Ilya
  Sutskever, T.~Lillicrap, M.~Leach, K.~Kavukcuoglu, T.~Graepel, and Demis
  Hassabis.
\newblock Mastering the game of go with deep neural networks and tree search.
\newblock \emph{Nature}, 529:\penalty0 484--489, 2016.


\bibitem[Sutton and Barto(1998)]{Sutton:1998}
Richard~S. Sutton and Andrew~G. Barto.
\newblock \emph{Introduction to Reinforcement Learning}.
\newblock MIT Press, Cambridge, MA, USA, 1st edition, 1998.
\newblock ISBN 0262193981.


\bibitem[Tao et~al.(2018)Tao, C{\^o}t{\'e}, Yuan, and Asri]{Tao:2018}
Ruo~Yu Tao, Marc-Alexandre C{\^o}t{\'e}, Xingdi Yuan, and Layla~El Asri.
\newblock Towards solving text-based games by producing adaptive action spaces.
\newblock \emph{ArXiv}, abs/1812.00855, 2018.

\bibitem[Trischler et~al.(2019)Trischler, C\^ot\'{e}, and Lima]{Trischler:2019}
Adam Trischler, Marc-Alexandre C\^ot\'{e}, and Pedro Lima.
\newblock First textworld problems, the competition: Using text-based games to
  advance capabilities of ai agents.
\newblock 2019.
\newblock URL
  \url{https://www.microsoft.com/en-us/research/blog/first-textworld-problems-the-competition}.


\bibitem[Urbanek et~al.(2019)Urbanek, Fan, Karamcheti, Jain, Humeau, Dinan,
  Rockt{\"a}schel, Kiela, Szlam, and Weston]{Urbanek:2019}
Jack Urbanek, Angela Fan, Siddharth Karamcheti, Saachi Jain, Samuel Humeau,
  Emily Dinan, Tim Rockt{\"a}schel, Douwe Kiela, Arthur Szlam, and J.~Weston.
\newblock Learning to speak and act in a fantasy text adventure game.
\newblock \emph{ArXiv}, abs/1903.03094, 2019.


\bibitem[Vaswani et~al.(2017)Vaswani, Shazeer, Parmar, Uszkoreit, Jones, Gomez,
  Kaiser, and Polosukhin]{Vaswani:2017}
Ashish Vaswani, Noam Shazeer, Niki Parmar, Jakob Uszkoreit, Llion Jones,
  Aidan~N Gomez, {\L}ukasz Kaiser, and Illia Polosukhin.
\newblock Attention is all you need.
\newblock \emph{Advances in neural information processing systems},
  30:\penalty0 5998--6008, 2017.

\bibitem[Xu et~al.(2020)Xu, Fang, Chen, Du, Zhou, and Zhang]{Xu:2020sha}
Yunqiu Xu, Meng Fang, Ling Chen, Yali Du, Joey~Tianyi Zhou, and Chengqi Zhang.
\newblock Deep reinforcement learning with stacked hierarchical attention for
  text-based games.
\newblock \emph{Advances in Neural Information Processing Systems}, 33, 2020.

\bibitem[{Xu} et~al.(2020){Xu}, {Chen}, {Fang}, {Wang}, and {Zhang}]{Xu:2020tf}
Y.~{Xu}, L.~{Chen}, M.~{Fang}, Y.~{Wang}, and C.~{Zhang}.
\newblock Deep reinforcement learning with transformers for text adventure
  games.
\newblock In \emph{2020 IEEE Conference on Games (CoG)}, pages 65--72, 2020.
\newblock \doi{10.1109/CoG47356.2020.9231622}.

\bibitem[Yao et~al.(2020)Yao, Rao, Hausknecht, and Narasimhan]{Yao:2020}
Shunyu Yao, Rohan Rao, Matthew Hausknecht, and Karthik Narasimhan.
\newblock Keep calm and explore: Language models for action generation in
  text-based games.
\newblock In \emph{Empirical Methods in Natural Language Processing (EMNLP)},
  2020.

\bibitem[Yin and May(2020)]{Yin:2020snn}
Xusen Yin and Jonathan May.
\newblock Zero-shot learning of text adventure games with sentence-level
  semantics.
\newblock \emph{arXiv preprint arXiv:2004.02986}, 2020.

\bibitem[Yin et~al.(2020)Yin, Weischedel, and May]{Yin:2020bert}
Xusen Yin, R.~Weischedel, and Jonathan May.
\newblock Learning to generalize for sequential decision making.
\newblock \emph{EMNLP}, 2020.


\bibitem[Yuan et~al.(2018)Yuan, C{\^o}t{\'e}, Sordoni, Laroche, des Combes,
  Hausknecht, and Trischler]{Yuan:2018}
Xingdi Yuan, Marc-Alexandre C{\^o}t{\'e}, Alessandro Sordoni, Romain Laroche,
  Remi~Tachet des Combes, Matthew Hausknecht, and Adam Trischler.
\newblock Counting to explore and generalize in text-based games.
\newblock \emph{35th International Conference on Machine Learning, Exploration
  in Reinforcement Learning Workshop}, 2018.

\bibitem[Yuan et~al.(2019)Yuan, C\^ot\'{e}, Fu, Lin, Pal, Bengio, and
  Trischler]{Yuan:2019}
Xingdi Yuan, Marc-Alexandre C\^ot\'{e}, Jie Fu, Zhouhan Lin, Christopher Pal,
  Yoshua Bengio, and Adam Trischler.
\newblock Interactive language learning by question answering.
\newblock 2019.


\bibitem[Zahavy et~al.(2018)Zahavy, Haroush, Merlis, Mankowitz, and
  Mannor]{Zahavy:2018}
Tom Zahavy, Matan Haroush, Nadav Merlis, Daniel~J Mankowitz, and Shie Mannor.
\newblock Learn what not to learn: Action elimination with deep reinforcement
  learning.
\newblock pages 3562--3573, 2018.


\bibitem[Zelinka et~al.(2019)Zelinka, Yuan, C{\^o}t{\'e}, Laroche, and
  Trischler]{Zelinka:2019}
Mikul{\'a} Zelinka, Xingdi Yuan, Marc-Alexandre C{\^o}t{\'e}, R.~Laroche,
  and Adam Trischler.
\newblock Building dynamic knowledge graphs from text-based games.
\newblock \emph{ArXiv}, abs/1910.09532, 2019.



\end{thebibliography}

\end{document}